\documentclass{aifrontiers}
\usepackage{aifrontiers}
\usepackage{booktabs}
\usepackage{array}
\usepackage{colortbl}
\usepackage{caption}
\usepackage{subcaption}
\usepackage{multirow}
\usepackage{float}
\usepackage{enumitem}
\usepackage{xspace}
\usepackage{makecell}

\usepackage[T1]{fontenc}    
\usepackage[utf8]{inputenc} 

\newlength\savewidth

\newcolumntype{x}[1]{>{\centering\arraybackslash}p{#1pt}}
\newcolumntype{y}[1]{>{\raggedright\arraybackslash}p{#1pt}}

\definecolor{baselinecolor}{gray}{0.93}

\begin{document}

\title{PrecisionCUA: Iterative Visual Refinement for Pixel-Precise Cursor Grounding in Code Editors}
\shorttitle{}
\author{
    Himangi Mittal$^{1,2*}$, Gaurav Mittal$^{1}$, Nelson Daniel Troncoso$^{1}$, Yu Hu$^{1}$\\
    {\normalsize $^{1}$Microsoft \quad $^{2}$Carnegie Mellon University}
}
\date{\today}

\renewcommand{\weblink}{}
\renewcommand{\foundrylink}{}
\renewcommand{\hflink}{}
\renewcommand{\ghlink}{}

\begin{abstract}
Computer Use Agents (CUAs) fundamentally rely on graphical user interface (GUI) grounding to translate language instructions into executable screen actions, but editing-level grounding in dense coding interfaces (such as VS Code and Cursor), where sub-pixel accuracy is required to interact with dense IDE elements, remains underexplored. Existing approaches typically rely on single-shot coordinate prediction, which lacks a mechanism for error correction and often fails in high-density interfaces. In this technical report, we conduct an empirical study of pixel-precise cursor localization in coding environments. Instead of a single-step execution, our agent engages in an iterative refinement process, utilizing visual feedback
from previous attempts to reach the target element. This closed-loop grounding mechanism allows the agent to self-correct displacement errors and adapt to dynamic UI changes. We evaluate our approach across Claude, Qwen, and GPT on a suite of complex coding benchmarks, demonstrating that multi-turn refinement significantly outperforms state-of-the-art single-shot models in both click precision and overall task success rate. Our results suggest that iterative visual reasoning is a critical component for the next generation of reliable
software engineering agents. Code: \url{https://github.com/microsoft/precision-cua-bench/tree/main}.   

\end{abstract}

\maketitle
\begingroup
\renewcommand{\thefootnote}{*}
\footnotetext{This work was done as Himangi Mittal's project at Microsoft CoreAI.}
\endgroup



\section{Introduction}

Computer Use Agents (CUAs)~\cite{wang2025opencua,agashe2025agent} are AI systems that interact with software interfaces in a human-like manner by perceiving the screen, interpreting natural-language instructions, and executing actions such as clicking, typing, selecting, and navigating across applications. They pave the way towards general-purpose digital task automation, with potential impact across productivity workflows, software operations, web navigation, customer support tooling, and developer assistance. Following rapid progress in large multimodal models and agentic reasoning pipelines, CUA-style interaction has emerged as a fast-growing frontier: both research and product efforts are moving quickly from static benchmarks toward real desktop and browser environments where reliability and precision directly affect utility.

A core capability underlying nearly every CUA action is GUI grounding~\cite{yang2025gta1,zhang2026tongui,gou2024navigating,wu2025gui,chen2026gui,tang2025gui,liu2025infigui}: mapping a user instruction to the correct spatial target on the screen. Before an agent can execute higher-level behavior, it must first localize \emph{where} to act. Even small localization errors propagate into failed clicks, incorrect edits, and broken multi-step trajectories, making robust grounding the bridge between language understanding and executable interface control.

Despite strong progress on general GUI benchmarks, current grounding performance is uneven across interaction types. Frontier models are increasingly reliable at coarse targets such as buttons, tabs, and icons, but remain substantially less reliable on editing-level actions that require fine-grained placement inside text-dense interfaces. The shortfall is qualitative: many models identify the right semantic region yet still miss the exact actionable pixel. The effect is amplified in dense IDE layouts, where line numbers, syntax-highlighted tokens, punctuation, and narrow cursor boundaries create visually crowded targets with minimal tolerance for error. Relative to button-click tasks, this editing-level regime remains underexplored in both evaluation design and capability analysis.

Consider an instruction such as: \emph{rename the function by placing the cursor between the characters ``n'' and ``a'' in \texttt{function}}. A prediction that lands even a few pixels to the left or right may target a neighboring token, insert text at the wrong boundary, or trigger an unintended edit despite being semantically close. Human users do not solve such tasks in a single move; instead, they make small visual corrections, repeatedly adjusting based on immediate feedback from the interface. We take this behavior as an analysis lens for model grounding and frame this work around the following study question: \emph{does explicit visual feedback help current frontier models correct grounding errors?}

To answer this question, we conduct a controlled comparison between two settings: (i) one-shot grounding, where the model predicts coordinates in a single attempt, and (ii) feedback-guided iterative grounding, where a subsequent attempt is made after showing a red-cross marker at the prior prediction. Both settings use the same task distribution, screenshots, instructions, and coordinate-extraction pipeline, isolating the effect of feedback from confounders such as data mixture or prompt format drift. The study dataset contains 5{,}390 annotated samples collected across two modern code editors that share the Electron + Monaco rendering stack (VS Code and Cursor) and across both Dark and Light themes, with coverage spanning line-, word-, and character-level grounding targets. Across evaluated models, the findings reveal a consistent but capacity-gated pattern: explicit feedback substantially improves grounding for frontier closed-source models, while smaller open-weight backbones gain little from the same protocol. Claude Opus 4.7 more than doubles its accuracy across five turns, whereas Qwen-3.5-9B and GPT-5.4-Pro remains within noise of zero throughout. A complementary comparison against specialized GUI grounding baselines and a lightweight in-domain finetune further shows that targeted task-specific supervision can surpass the strongest zero-shot multi-turn configuration.

In summary, this technical report makes three contributions: (i) a capability-oriented benchmark focused on editing-level GUI grounding in dense code editors, where pixel-precise cursor placement is required; (ii) a controlled analysis of explicit visual feedback that compares one-shot and feedback-guided evaluation under matched task and parsing conditions; and (iii) empirical insights into how correction dynamics vary across frontier models, prompting strategies, editor and theme conditions, and lightweight in-domain finetuning.

\vspace{-1em}
\section{Related Work}
\begin{figure}[t]
\centering
\includegraphics[width=0.95\linewidth]{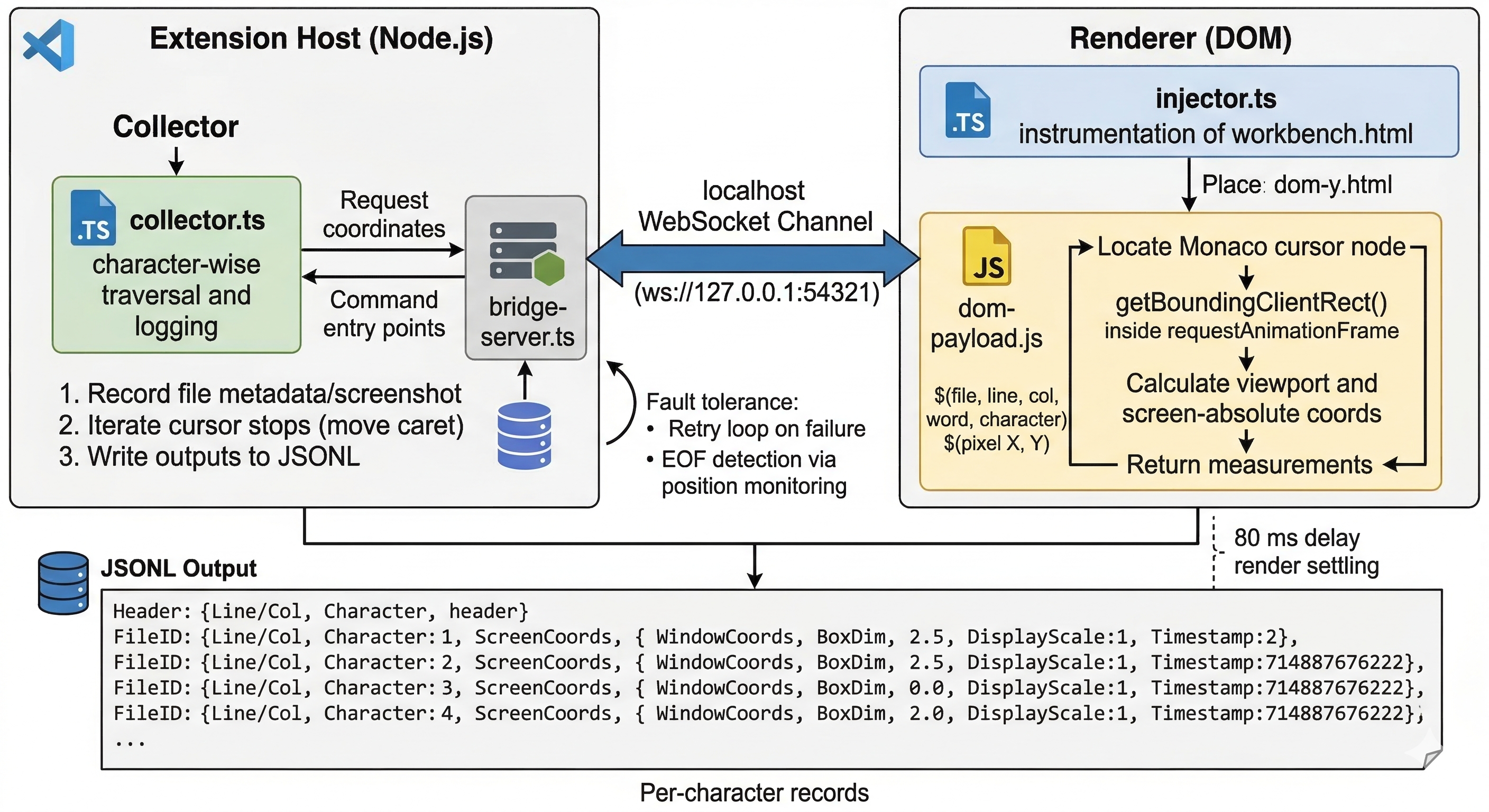}
\caption{Data collection system overview. A process-separated architecture connects the editor extension host (VS Code or Cursor) with the renderer DOM through a localhost WebSocket channel: the extension iterates over symbolic cursor stops and the renderer-side payload measures the corresponding pixel geometry via \texttt{getBoundingClientRect()} inside \texttt{requestAnimationFrame}. The synchronized records are written to JSONL with screen-absolute coordinates and editor metadata. Because both editors share the Electron+Monaco stack, the same pipeline produces directly comparable records across them.}
\label{fig:data-collection-overview}
\vspace{-10pt}
\end{figure}
\textbf{Computer-Use Agents.} Computer-Use Agents (CUAs) are a specialized class of multimodal systems designed to perceive and interact with digital environments through graphical user interfaces (GUIs) in a manner that mirrors human operation. The architectural paradigm has undergone a fundamental shift from specialized, task-specific models~\cite{humphreys2022data,shi2017world} to general-purpose foundation agents~\citep{deng2023mind2web,hong2024cogagent,zhang2025appagent,he2024webvoyager}. Task-specific models, while effective in narrow domains, often struggle to generalize across different operating systems or complex, non-stationary applications. Foundation agents leverage the pre-trained reasoning and spatial understanding of frontier models~\cite{hong2024cogagent}~(e.g., GPT-4V and Claude 3.5) to operate in a training-free or few-shot manner. Recent work has explored the different modalities associated with CUAs. Computer-Use Agents primarily perceive their environment through two distinct modalities: structured text and raw visual input. Text-based approaches~\cite{deng2023mind2web} rely on underlying metadata such as HTML DOM trees or accessibility APIs to interpret the UI. However, these agents are often brittle when faced with dynamic web content or applications. In contrast, vision-based (screenshot-only) agents~\cite{hong2024cogagent} process raw pixel data directly. This helps models like GPT-4V and CogAgent achieve greater robustness across diverse interfaces without requiring access to the underlying source code. However, vision-only models still encounter significant challenges in high-precision coordinate prediction. The ``precision gap'' in mapping visual intent to exact pixel coordinates, especially in dense environments like IDEs, remains a fundamental bottleneck that our work seeks to address through iterative visual refinement.

\textbf{GUI Grounding} Recent GUI-grounding research has progressed from coarse element localization toward more general and scalable visual grounding for agentic interaction. Early work such as SeeClick demonstrates that large multimodal models can ground instructions to UI targets across heterogeneous screens~\cite{seeclick}, while newer benchmarks such as ScreenSpot-Pro emphasize higher-resolution and professional-use settings where fine localization becomes more demanding~\cite{screenspotpro}. Parallel lines of work focus on improving grounding mechanisms and representations: test-time scaling and search-based strategies improve inference-time correction dynamics~\cite{yang2025gta1}, coordinate-free formulations reduce direct dependence on absolute click regression~\cite{wu2025gui}, and tool-augmented or reward-model-based approaches improve training signals for difficult grounding cases~\cite{chen2026gui,tang2025gui}. More recent efforts also explore large-scale trajectory supervision and deliberative multimodal reasoning for broader GUI generalization~\cite{zhang2026tongui,liu2025infigui}. Despite this progress, editing-level grounding in dense coding interfaces remains comparatively undercharacterized: many systems report strong performance on larger clickable elements, but precise cursor placement between nearby characters is still fragile. Our work complements prior method-driven advances by providing a focused empirical study of this precision regime and analyzing how explicit visual feedback affects correction behavior across frontier models.
\section{Methodology}

\vspace{-3pt}
\subsection{Problem Formulation}
We formulate GUI grounding as a multi-turn iterative refinement approach. Given a natural-language instruction $I$ and an initial visual state $S_1$, the goal is to predict the target pixel location $C^*=(x^*,y^*)$. Rather than treating grounding as a single-shot prediction, we model it as a $T$-step refinement process that progressively reduces localization error. To supervise this objective at pixel precision, we next construct a dedicated data collection pipeline that aligns symbolic cursor indices with renderer-space coordinates.

\vspace{-3pt}
\subsection{Data Collection}
\label{sec:data-collection}
To realize the formulation above, we build a purpose-designed data-collection pipeline (\texttt{vscode-cursor-coords}) that runs as an editor extension and generates ground-truth mappings from symbolic cursor states (\texttt{file}, \allowbreak\texttt{line}, \allowbreak\texttt{col}, 
\allowbreak\texttt{word}, \allowbreak\texttt{character}) to measured Monaco pixel coordinates. Because both VS Code and Cursor are built on the same Electron + Monaco renderer stack, a single extension and DOM-payload injection path supports both editors\footnote{Throughout this paper, ``Cursor'' (capitalized) refers to the AI-IDE \href{https://www.cursor.com/}{cursor.com}; we use ``text cursor'' or ``caret'' for the editor insertion point.}, yielding aligned data across two of the most common modern coding environments. The central systems challenge is process separation: cursor indices are exposed in the extension host, whereas pixel geometry is observable only in the renderer DOM. We bridge this gap with a localhost WebSocket channel connecting a Node.js backend to an injected renderer-side payload script that measures Monaco caret bounding boxes via \texttt{getBoundingClientRect()} inside \texttt{requestAnimationFrame}, and converts viewport-relative CSS coordinates to screen-absolute coordinates. The collector iterates one cursor step at a time, including end-of-line stops, and writes per-character JSONL records that store both coordinate frames, cursor box geometry, and \texttt{devicePixelRatio} for deterministic replay across heterogeneous monitor DPI settings. Figure~\ref{fig:data-collection-overview} illustrates the end-to-end architecture; full system, schema, and robustness details are deferred to Appendix~\ref{app:data-collection}.

\vspace{-3pt}
\subsection{Multi-Turn Iterative Grounding Loop}
Using the collected supervision, our model performs iterative grounding with one initial prediction followed by $T-1$ refinement turns.

\textbf{Turn 1: Initial Prediction} The agent receives the instruction $I$ and the raw screenshot $S_1$. The model $\pi_{\theta}$ predicts the first coordinate:
\begin{equation}
    C_1 = \pi_{\theta}(I, S_1)
\end{equation}

\textbf{Turn $t > 1$: Visual Feedback Refinement.} For subsequent turns, the agent is provided with the visual feedback of its previous error. We define a visual marking function $V(S_t, C_{t-1})$ that renders a red cross-hair at coordinate $C_{t-1}$ on the current screenshot $S_t$. The agent then predicts the next coordinate $C_t$ based on the instruction, the marked screenshot, and the numerical value of the previous coordinate:
\begin{equation}
    C_t = \pi_{\theta}(I, V(S_t, C_{t-1}), C_{t-1})
\end{equation}
This design exposes the model to explicit spatial error signals between successive attempts, enabling a form of visual servoing in latent space that progressively sharpens localization.

\vspace{-3pt}
\subsection{Prompting}
To build a robust multi-turn approach that can leverage information from previous turns, we explore prompting-based strategies along two orthogonal controls: a \emph{system prompt variant} (global behavioral policy for all turns) and a \emph{feedback template variant} (turn-conditioned correction cue used after misses). The system prompt defines \emph{how} the model should look and reason; the feedback template defines \emph{how} the model should react to failure signals.

\textbf{System prompts.} The harness defines seven named system prompts: \texttt{baseline}, \texttt{baseline\_cot}, \texttt{cursor\_aware}, \texttt{step\_by\_step}, \texttt{minimal}, \texttt{visual\_anchor}, and a \texttt{custom} placeholder. All are parameterized by image dimensions and share three constraints: acknowledge red-cross feedback if present, localize in pixel coordinates, and end with a bare \texttt{(x,y)} pair for deterministic parsing. The set spans a deliberate spectrum from low-structure generic prompts (\texttt{minimal}, \texttt{baseline}) to explicit chain-of-thought (\texttt{baseline\_cot}, \texttt{step\_by\_step}) and task-specialized priors (\texttt{cursor\_aware}, \texttt{visual\_anchor}), enabling controlled ablations of reasoning granularity, domain specialization, and anchor-based spatial strategy. The four variants used in Section~\ref{sec:prompt-design} (Baseline, Baseline-CoT, Cursor-Aware, Visual-Anchor) are the most informative subset; the full prompt text for all seven variants is reproduced in Appendix~\ref{app:system-prompts}.

\textbf{Feedback templates.} Between turns, the harness selects one of two textual feedback templates, \texttt{baseline} or \texttt{spatial}, both paired with a red-cross overlay at the prior prediction. \texttt{baseline} simply states that the previous answer was incorrect, whereas \texttt{spatial} explicitly asks the model to reason about where the cross lies relative to the target and adjust directionally. Feedback in this protocol is therefore a synchronized multimodal signal: the textual template encodes correction intent and the overlay provides the geometric evidence. All experiments in Section~\ref{sec:multi-turn} use the \texttt{spatial} template; full template text and minor harness details (non-cumulative overlay redraw, no full-history recap) are given in Appendix~\ref{app:feedback-templates}.

\section{Experiments}

\subsection{Implementation Details}
\label{sec:impl-details}

\textbf{Models.} We evaluate the multi-turn refinement framework on two models that span the open / closed-source and small / frontier axes: \textbf{Qwen-3.5-9B} (open-weight, 9B parameters), \textbf{Anthropic Claude Opus 4.7} (closed-source frontier), and \textbf{GPT-5.4-Pro}~(closed-source). This pairing lets us isolate the contribution of the iterative protocol from raw model capacity. Section~\ref{sec:finetune-baselines} additionally compares against two specialized GUI grounding baselines (GUI-Actor and GUI-G\textsuperscript{2}) and a lightweight in-domain finetune of GTA1, evaluated under matched conditions on the full benchmark.

\textbf{Data.} Evaluation is performed on our text-cursor benchmark over two modern code editors, VS Code and Cursor. The full release contains 5{,}390 annotated samples, balanced across editors (2{,}655 VS Code / 2{,}735 Cursor), themes (2{,}697 Dark / 2{,}693 Light), and target granularities (3{,}518 character / 1{,}092 word / 780 line). All reported model results in Tables~\ref{tbl:main-results}--\ref{tbl:claude-ide-theme} are computed on the full 5{,}390-sample benchmark, with every model and prompt condition evaluated on the same set of samples to make per-cell comparisons paired. All screenshots are normalized to a standard $1280 \times 1024$ resolution before being passed to the model; cross-editor comparisons remain valid because both editors share the Monaco renderer geometry.

\textbf{Visual feedback rendering.} In each refinement turn $t > 1$, the visual marking function $V(S_t, C_{t-1})$ renders a semi-transparent red cross-hair (hex \texttt{\#FF0000}) centered on the previous prediction $(x_{t-1}, y_{t-1})$. The cross-hair spans $5\%$ of the image width and height to remain visible without occluding the underlying UI. The numerical coordinates are also appended to the text prompt as \texttt{Last attempt: [x, y]}.

\textbf{Inference harness.} The all-turns harness consumes JSONL samples containing an image path, a natural-language instruction, and a ground-truth bounding box in normalized $[0,1000]$ coordinates, which are rescaled to pixel space at runtime via image-specific width/height ratios. At turn 1, the system prompt (parameterized by image dimensions) and user instruction are sent to the model backend (Foundry Claude, GPT-5.4-Pro, or an OpenAI-compatible Qwen endpoint); responses are parsed with a coordinate regex, and the final coordinate pair in the output is taken as the model's decision. For $t>1$, if the previous prediction missed, the harness draws the red cross marker on a clean copy of the original screenshot, appends the prior assistant turn to the dialogue history, and injects the spatial-feedback template before re-querying. The loop uses early stopping on first hit and records parse failures alongside both point-to-bounding-box and point-to-center distances; degenerate point-like targets are handled via a tolerance-based hit rule. Aggregation follows cumulative multi-turn semantics: a hit at turn $T$ is carried forward as a hit for turns $T{+}1\ldots N$, so reported ``accuracy at turn $T$'' is success within at most $T$ attempts. Unresolved samples propagate their final-turn miss distance for consistent cross-turn distance reporting. Open-weight inference (Qwen-3.5-9B) and GTA1 finetuning each use a single node of $8\times$ NVIDIA A100 GPUs, while closed-source models (Claude Opus 4.7, GPT-5.4) are accessed through their respective hosted APIs.

\subsection{Multi-Turn Refinement}
\label{sec:multi-turn}

Tables~\ref{tbl:main-results} and~\ref{tbl:elementwise-results} report cumulative localization accuracy, point-to-bounding-box distance ($D_{\text{bb}}$), point-to-center distance ($D_{\text{c}}$), and per-granularity accuracy across five refinement turns ($T_1$ through $T_5$). Three observations are consistent across the four prompt variants.

\textbf{Iterative feedback is effective for capable backbones.} Claude Opus 4.7 improves substantially under refinement: accuracy rises from approximately 21\% at $T_1$ to between 44.6\% and 48.1\% at $T_5$, a gain of roughly 25 percentage points. The Baseline-CoT prompt yields the largest absolute improvement (22.8\% to 48.1\%), and the relative gain is concentrated in turns 2 and 3, after which the curve flattens. This pattern indicates that the red-cross marker and the textual ``last attempt'' coordinate provide a usable correction signal, rather than simply increasing inference compute. GPT-5.4-Pro gains even more from refinement under text-only prompts (Baseline 13.5\%$\to$41.0\%, Baseline-CoT 11.1\%$\to$41.2\%), with lower residual $D_{\text{bb}}$ than Claude (255.8 vs.\ 411.8 px at $T_5$) and higher $T_5$ line (70.2\%) and word (62.5\%) accuracy (Table~\ref{tbl:elementwise-results}). It is far more prompt-sensitive, however: Cursor-Aware and Visual-Anchor cap $T_5$ at 16.9\% and 18.5\%. The same protocol does not rescue Qwen-3.5-9B, whose accuracy remains in the 1.4\%--2.6\% range across all turns and prompts. Iterative visual grounding therefore appears to be capacity-gated: the base model must already encode a usable spatial prior before feedback can be exploited, and for GPT-5.4-Pro the prompt scaffolding itself can suppress that prior. 

\textbf{Distance metrics increase with turns under cumulative semantics.} For Claude Opus 4.7, $D_{\text{bb}}$ grows monotonically with $T$ in every prompt setting (e.g., Baseline: 292.4 px at $T_1$ to 411.8 px at $T_5$). This trend is not a regression in spatial precision but a direct consequence of the cumulative aggregation rule defined in Section~\ref{sec:impl-details}. Once an easy sample is solved at turn $T$, it is removed from the residual mass that drives the distance statistic; subsequent turns therefore aggregate over a progressively harder, never-corrected subset whose miss distances are large. We accordingly read $D_{\text{bb}}$ and $D_{\text{c}}$ as conditional measures of the residual difficulty and recommend interpreting them jointly with cumulative accuracy rather than as independent precision metrics.

\textbf{Granularity defines a clear difficulty ordering.} Element-wise accuracy in Table~\ref{tbl:elementwise-results} establishes a stable ordering Line $>$ Word $>$ Character for Claude Opus 4.7 across every prompt and turn. Line accuracy saturates near 60\% by $T_2$ and grows only marginally thereafter (e.g., Baseline-CoT: 57.2\% at $T_1$ to 62.4\% at $T_5$), suggesting that line-level grounding is largely a turn-1 phenomenon. Word accuracy benefits more from refinement (Baseline-CoT: 42.0\% to 58.2\%), with most of the gain occurring between $T_1$ and $T_2$. Character accuracy is the dominant residual: even the best multi-turn configuration reaches only 41.8\% at $T_5$, compared to 9.1\% at $T_1$. The gap between line and character accuracy widens with turns, indicating that successive feedback rounds disproportionately resolve coarse-grained errors while sub-glyph localization remains an open challenge. GPT-5.4-Pro reproduces the same Line $>$ Word $>$ Character ordering: under Baseline-CoT, line accuracy rises from 31.7\% at $T_1$ to 60.6\% at $T_2$ and 70.2\% at $T_5$, word from 20.9\% to 62.5\%, while character climbs only to 28.1\%; Cursor-Aware and Visual-Anchor suppress all three granularities below 33\% at $T_5$. Qwen-3.5-9B exhibits near-zero character and word accuracy at all turns, confirming that without sufficient base capability, finer granularities are inaccessible regardless of prompt or refinement budget.


\begin{table}[t]
\centering
\scriptsize
\setlength{\tabcolsep}{2.5pt}
\renewcommand{\arraystretch}{1.05}
\resizebox{\textwidth}{!}{%
\begin{tabular}{p{1.1cm}l|ccccc|ccccc|ccccc}
\hline
 & & \multicolumn{5}{c|}{\textbf{Accuracy (\%)} $\uparrow$} & \multicolumn{5}{c|}{\textbf{D$_{\text{bb}}$ (px)} $\downarrow$} & \multicolumn{5}{c}{\textbf{D$_{\text{c}}$ (px)} $\downarrow$} \\
\textbf{Prompt} & \textbf{Model} & T1 & T2 & T3 & T4 & T5 & T1 & T2 & T3 & T4 & T5 & T1 & T2 & T3 & T4 & T5 \\
\hline
\multirow{2}{1.1cm}{\textbf{Baseline}}
 & Qwen 3.5 9B     &   1.5 &   1.6 &   1.7 &   1.7 &   1.8 & 272.0 & 266.6 & 266.3 & 265.2 & 264.9 & 284.0 & 278.0 & 277.5 & 276.5 & 276.1 \\
  & GPT-5.4-Pro &  13.5 &  25.3 &  32.2 &  37.0 &  41.0 & 185.2 & 216.4 & 220.2 & 235.2 & 255.8 & 195.3 & 224.4 & 225.5 & 240.8 & 261.4 \\
 & Claude Opus 4.7 &  21.0 &  31.0 &  37.5 &  42.0 &  45.4 & 292.4 & 327.6 & 360.1 & 388.1 & 411.8 & 301.0 & 336.7 & 368.4 & 396.4 & 420.3 \\
\hline
\multirow{2}{1.1cm}{\textbf{Baseline CoT}}
& Qwen 3.5 9B     &   1.4 &   1.9 &   2.1 &   2.3 &   2.6 & 450.3 & 437.4 & 451.0 & 445.5 & 445.1 & 459.5 & 444.0 & 457.2 & 451.4 & 450.2 \\
 & GPT-5.4-Pro &  11.1 &  25.8 &  33.0 &  37.5 &  41.2 & 181.5 & 230.2 & 234.6 & 252.0 & 257.5 & 191.6 & 237.5 & 240.6 & 258.2 & 262.8 \\
& Claude Opus 4.7 &  22.8 &  32.8 &  39.1 &  44.0 &  48.1 & 296.9 & 335.0 & 368.8 & 400.5 & 431.7 & 305.4 & 343.0 & 377.6 & 409.1 & 441.2 \\
\hline
\multirow{2}{1.1cm}{\textbf{Cursor-Aware}}
& Qwen 3.5 9B     &   1.2 &   1.6 &   1.7 &   1.9 &   1.9 & 443.5 & 433.5 & 426.7 & 427.8 & 428.6 & 449.6 & 438.4 & 431.5 & 432.6 & 433.5 \\
   & GPT-5.4-Pro &   1.5 &   4.8 &  11.0 &  14.5 &  16.9 & 189.5 & 178.1 & 198.7 & 202.9 & 214.7 & 205.5 & 190.5 & 210.0 & 212.8 & 223.2 \\
& Claude Opus 4.7 &  20.9 &  30.7 &  36.8 &  41.7 &  44.9 & 289.1 & 323.4 & 355.0 & 384.2 & 406.4 & 297.4 & 330.4 & 363.5 & 392.8 & 415.2 \\
\hline
\multirow{2}{1.1cm}{\textbf{Visual-Anchor}}
& Qwen 3.5 9B     &   1.3 &   1.5 &   1.7 &   1.8 &   1.9 & 449.1 & 427.0 & 426.9 & 427.9 & 428.9 & 457.0 & 432.7 & 432.7 & 433.5 & 434.7 \\
& GPT-5.4-Pro &   3.1 &   6.9 &  12.2 &  15.6 &  18.5 & 186.2 & 181.2 & 194.6 & 198.5 & 207.8 & 202.0 & 191.5 & 203.8 & 207.5 & 214.8 \\
& Claude Opus 4.7 &  21.8 &  31.3 &  36.9 &  41.3 &  44.6 & 339.8 & 381.8 & 415.2 & 445.6 & 472.1 & 348.9 & 390.0 & 424.2 & 454.2 & 480.9 \\
\hline
\end{tabular}%
}
\caption{\textbf{Multi-turn grounding results.} Accuracy (\%), point-to-bounding-box distance (D$_{\text{bb}}$, px) and point-to-center distance (D$_{\text{c}}$, px) across five refinement turns (T1--T5) for each of the four system-prompt variants and the three evaluated models (Qwen-3.5-9B, Claude Opus 4.7, and GPT-5.4-Pro). Each cell is averaged over the full 5{,}390-sample benchmark, spanning both editors (VS Code, Cursor) and both themes (Dark, Light). Higher is better for accuracy; lower is better for distances. Distances are reported under cumulative semantics and therefore aggregate over the residual unresolved subset at each turn (Section~\ref{sec:multi-turn}).}
\label{tbl:main-results}

\label{tbl:main-results}
\end{table}

\begin{table}[t]
\centering
\scriptsize
\setlength{\tabcolsep}{2.5pt}
\renewcommand{\arraystretch}{1.05}
\resizebox{\textwidth}{!}{%
\begin{tabular}{p{1.1cm}l|ccccc|ccccc|ccccc}
\hline
 & & \multicolumn{5}{c|}{\textbf{Character Acc. (\%)} $\uparrow$} & \multicolumn{5}{c|}{\textbf{Word Acc. (\%)} $\uparrow$} & \multicolumn{5}{c}{\textbf{Line Acc. (\%)} $\uparrow$} \\
\textbf{Prompt} & \textbf{Model} & T1 & T2 & T3 & T4 & T5 & T1 & T2 & T3 & T4 & T5 & T1 & T2 & T3 & T4 & T5 \\
\hline
\multirow{2}{1.1cm}{\textbf{Baseline}}
 & Qwen 3.5 9B     &   0.0 &   0.1 &   0.1 &   0.1 &   0.1 &   0.4 &   0.5 &   0.5 &   0.6 &   0.8 &   9.5 &  10.0 &  10.4 &  10.5 &  10.9 \\
  & GPT-5.4-Pro &   3.5 &   9.3 &  15.8 &  21.9 &  27.2 &  28.5 &  49.8 &  59.0 &  62.1 &  64.3 &  36.9 &  62.4 &  67.7 &  69.1 &  70.4 \\
 & Claude Opus 4.7 &   6.9 &  16.9 &  26.1 &  32.6 &  37.4 &  41.0 &  56.1 &  57.6 &  58.4 &  59.0 &  56.9 &  59.3 &  60.7 &  61.7 &  62.3 \\
\hline
\multirow{2}{1.1cm}{\textbf{Baseline CoT}}
 & Qwen 3.5 9B     &   0.0 &   0.0 &   0.1 &   0.2 &   0.3 &   0.0 &   0.0 &   0.0 &   0.0 &   0.4 &   9.7 &  13.3 &  14.4 &  14.9 &  16.4 \\
  & GPT-5.4-Pro &   3.5 &  11.0 &  18.3 &  23.7 &  28.1 &  20.9 &  48.3 &  56.7 &  59.8 &  62.5 &  31.7 &  60.6 &  65.9 &  68.3 &  70.2 \\
 & Claude Opus 4.7 &   9.1 &  19.7 &  28.5 &  35.6 &  41.8 &  42.0 &  55.7 &  57.6 &  58.0 &  58.2 &  57.2 &  59.9 &  61.2 &  62.2 &  62.4 \\
\hline
\multirow{2}{1.1cm}{\textbf{Cursor-Aware}}
& Qwen 3.5 9B     &   0.1 &   0.1 &   0.2 &   0.2 &   0.2 &   0.5 &   0.7 &   0.7 &   1.1 &   1.3 &   7.4 &  10.0 &  10.0 &  10.5 &  10.5 \\
  & GPT-5.4-Pro &   0.1 &   1.6 &   5.2 &   7.6 &   9.4 &   4.3 &   9.5 &  20.9 &  26.2 &  29.5 &   4.3 &  12.6 &  23.2 &  28.7 &  32.5 \\
& Claude Opus 4.7 &   7.7 &  17.1 &  25.0 &  32.2 &  36.9 &  39.7 &  53.8 &  57.6 &  58.2 &  58.3 &  54.2 &  60.1 &  60.7 &  61.4 &  62.1 \\
\hline
\multirow{2}{1.1cm}{\textbf{Visual-Anchor}}
& Qwen 3.5 9B     &   0.2 &   0.2 &   0.2 &   0.2 &   0.2 &   0.4 &   0.4 &   0.9 &   1.1 &   1.1 &   7.7 &   9.5 &   9.7 &  10.5 &  10.8 \\
  & GPT-5.4-Pro &   0.3 &   2.6 &   6.6 &   9.7 &  12.8 &   8.2 &  14.0 &  21.9 &  27.4 &  30.6 &   8.4 &  16.4 &  23.6 &  25.8 &  27.2 \\
& Claude Opus 4.7 &   7.5 &  17.6 &  25.6 &  32.3 &  37.5 &  39.5 &  52.6 &  54.3 &  54.7 &  54.8 &  55.6 &  58.2 &  59.1 &  59.5 &  59.8 \\
\hline
\end{tabular}%
}
\caption{\textbf{Multi-turn element-wise grounding accuracy.} Accuracy (\%) decomposed by target granularity (character, word, and line) across five refinement turns (T1--T5) for the four system-prompt variants and the three evaluated models (Qwen-3.5-9B, Claude Opus 4.7, and GPT-5.4-Pro). Computed on the full 5{,}390-sample benchmark (3{,}518 character, 1{,}092 word, 780 line), matching Table~\ref{tbl:main-results}. Higher is better.}
\label{tbl:elementwise-results}
\end{table}


\begin{table}[t]
\centering
\scriptsize
\setlength{\tabcolsep}{2.5pt}
\renewcommand{\arraystretch}{1.05}
\resizebox{\textwidth}{!}{%
\begin{tabular}{p{1.1cm}ll|ccccc|ccccc|ccccc}
\hline
\multicolumn{18}{c}{\textbf{Overall localization} \quad (Accuracy $\uparrow$, D$_{\text{bb}}$ $\downarrow$, D$_{\text{c}}$ $\downarrow$)} \\
\hline
 & & & \multicolumn{5}{c|}{\textbf{Accuracy (\%)} $\uparrow$} & \multicolumn{5}{c|}{\textbf{D$_{\text{bb}}$ (px)} $\downarrow$} & \multicolumn{5}{c}{\textbf{D$_{\text{c}}$ (px)} $\downarrow$} \\
\textbf{Prompt} & \textbf{IDE} & \textbf{Theme} & T1 & T2 & T3 & T4 & T5 & T1 & T2 & T3 & T4 & T5 & T1 & T2 & T3 & T4 & T5 \\
\hline
\multirow{4}{1.1cm}{\textbf{Baseline}}
 & \multirow{2}{*}{VS Code} & Dark  & 34.6 & 51.1 & 61.4 & 67.9 & 73.1 & 23.0 & 11.7 & 12.3 & 13.9 & 16.3 & 25.4 & 13.5 & 14.2 & 16.1 & 18.4 \\
 &                          & Light & 35.4 & 49.3 & 59.5 & 66.2 & 71.1 & 23.0 & 13.1 & 12.5 & 14.4 & 16.4 & 25.8 & 15.3 & 14.7 & 16.5 & 18.7 \\
 & \multirow{2}{*}{Cursor} & Dark  & 6.9 & 11.0 & 14.3 & 17.1 & 19.0 & 475.3 & 495.8 & 513.5 & 531.4 & 543.5 & 487.5 & 509.2 & 524.0 & 540.8 & 553.0 \\
 &                          & Light & 8.3 & 13.8 & 16.2 & 18.7 & 20.2 & 475.5 & 505.7 & 520.0 & 534.6 & 545.1 & 488.6 & 518.3 & 531.9 & 546.7 & 556.6 \\
\hline
\multirow{4}{1.1cm}{\textbf{Baseline CoT}}
 & \multirow{2}{*}{VS Code} & Dark  & 36.9 & 54.3 & 63.9 & 71.7 & 77.4 & 16.7 & 10.8 & 13.0 & 14.4 & 17.1 & 19.3 & 12.8 & 15.2 & 16.9 & 19.7 \\
 &                          & Light & 38.3 & 52.1 & 60.7 & 68.8 & 75.0 & 18.4 & 10.6 & 11.5 & 12.7 & 14.3 & 21.1 & 12.7 & 13.7 & 14.9 & 17.0 \\
 & \multirow{2}{*}{Cursor} & Dark  & 7.1 & 12.2 & 16.4 & 18.3 & 20.5 & 476.9 & 500.6 & 525.5 & 537.1 & 551.7 & 488.9 & 511.9 & 537.0 & 547.1 & 563.3 \\
 &                          & Light & 9.7 & 14.3 & 17.1 & 19.2 & 21.6 & 483.8 & 506.8 & 523.6 & 537.4 & 553.7 & 496.6 & 517.9 & 535.3 & 549.2 & 565.1 \\
\hline
\multirow{4}{1.1cm}{\textbf{Cursor-Aware}}
 & \multirow{2}{*}{VS Code} & Dark  & 33.0 & 49.0 & 60.7 & 69.4 & 74.0 & 18.9 & 11.4 & 12.9 & 15.8 & 17.1 & 23.2 & 14.4 & 15.8 & 19.1 & 20.8 \\
 &                          & Light & 35.1 & 49.0 & 56.6 & 64.0 & 68.3 & 19.4 & 11.2 & 12.7 & 13.2 & 14.1 & 23.8 & 13.8 & 15.2 & 15.8 & 16.6 \\
 & \multirow{2}{*}{Cursor} & Dark  & 7.7 & 11.7 & 14.3 & 16.8 & 18.9 & 478.6 & 496.9 & 513.6 & 528.1 & 541.9 & 489.1 & 506.8 & 524.4 & 538.7 & 552.4 \\
 &                          & Light & 8.6 & 14.3 & 16.8 & 18.2 & 20.0 & 475.4 & 504.7 & 521.8 & 530.0 & 542.6 & 487.1 & 513.7 & 533.7 & 541.3 & 553.8 \\
\hline
\multirow{4}{1.1cm}{\textbf{Visual-Anchor}}
 & \multirow{2}{*}{VS Code} & Dark  & 35.5 & 52.2 & 62.8 & 70.6 & 76.7 & 19.4 & 11.5 & 13.4 & 14.7 & 17.6 & 22.2 & 13.6 & 15.7 & 17.1 & 20.2 \\
 &                          & Light & 60.0 & 76.7 & 82.8 & 86.9 & 90.2 & 18.4 & 13.2 & 15.4 & 18.3 & 22.8 & 21.8 & 14.9 & 17.1 & 20.0 & 24.7 \\
 & \multirow{2}{*}{Cursor} & Dark  & 7.2 & 11.6 & 14.5 & 17.3 & 19.3 & 477.1 & 496.9 & 513.4 & 530.8 & 543.0 & 488.4 & 506.9 & 523.9 & 540.6 & 552.8 \\
 &                          & Light & 8.3 & 13.2 & 16.5 & 19.2 & 21.4 & 473.7 & 500.4 & 520.1 & 536.8 & 551.8 & 485.8 & 510.8 & 530.9 & 547.0 & 561.7 \\
\hline
\multicolumn{18}{c}{} \\[-1.6ex]
\hline
\multicolumn{18}{c}{\textbf{Element-wise accuracy} \quad (Character / Word / Line, \% $\uparrow$)} \\
\hline
 & & & \multicolumn{5}{c|}{\textbf{Character Acc. (\%)} $\uparrow$} & \multicolumn{5}{c|}{\textbf{Word Acc. (\%)} $\uparrow$} & \multicolumn{5}{c}{\textbf{Line Acc. (\%)} $\uparrow$} \\
\textbf{Prompt} & \textbf{IDE} & \textbf{Theme} & T1 & T2 & T3 & T4 & T5 & T1 & T2 & T3 & T4 & T5 & T1 & T2 & T3 & T4 & T5 \\
\hline
\multirow{4}{1.1cm}{\textbf{Baseline}}
 & \multirow{2}{*}{VS Code} & Dark  & 10.3 & 27.3 & 42.8 & 52.8 & 60.5 & 66.2 & 92.9 & 94.4 & 94.4 & 95.5 & 96.9 & 97.4 & 97.4 & 97.4 & 97.4 \\
 &                          & Light & 12.0 & 26.4 & 40.9 & 50.4 & 57.7 & 67.6 & 88.6 & 91.5 & 93.8 & 94.5 & 93.8 & 95.4 & 96.9 & 97.4 & 97.4 \\
 & \multirow{2}{*}{Cursor} & Dark  & 2.3 & 6.5 & 10.5 & 14.2 & 16.9 & 13.1 & 18.6 & 19.3 & 20.1 & 20.1 & 19.1 & 21.6 & 24.7 & 26.3 & 27.3 \\
 &                          & Light & 3.2 & 8.5 & 11.7 & 14.8 & 16.6 & 17.9 & 25.2 & 25.9 & 26.3 & 27.0 & 17.9 & 22.6 & 23.6 & 25.6 & 27.2 \\
\hline
\multirow{4}{1.1cm}{\textbf{Baseline CoT}}
 & \multirow{2}{*}{VS Code} & Dark  & 14.1 & 33.5 & 47.3 & 59.4 & 68.0 & 66.5 & 89.5 & 92.1 & 92.1 & 92.9 & 95.9 & 96.9 & 97.4 & 97.4 & 97.4 \\
 &                          & Light & 15.6 & 30.3 & 42.4 & 54.2 & 63.8 & 69.5 & 89.3 & 92.6 & 94.1 & 94.1 & 94.9 & 96.4 & 96.9 & 97.4 & 97.4 \\
 & \multirow{2}{*}{Cursor} & Dark  & 2.6 & 8.0 & 13.7 & 16.0 & 19.1 & 13.5 & 19.0 & 19.7 & 20.1 & 20.1 & 19.0 & 22.1 & 24.6 & 26.7 & 27.2 \\
 &                          & Light & 4.8 & 8.5 & 12.2 & 15.1 & 18.6 & 19.3 & 26.3 & 27.0 & 27.0 & 27.0 & 19.0 & 24.1 & 25.6 & 27.2 & 27.7 \\
\hline
\multirow{4}{1.1cm}{\textbf{Cursor-Aware}}
 & \multirow{2}{*}{VS Code} & Dark  & 10.8 & 27.9 & 43.0 & 55.9 & 63.1 & 62.6 & 83.0 & 91.9 & 93.3 & 93.3 & 89.7 & 94.9 & 95.4 & 95.4 & 95.4 \\
 &                          & Light & 14.1 & 26.0 & 36.4 & 47.3 & 53.7 & 66.2 & 89.7 & 93.0 & 93.8 & 93.8 & 84.1 & 93.8 & 94.4 & 95.9 & 96.9 \\
 & \multirow{2}{*}{Cursor} & Dark  & 2.6 & 6.9 & 10.2 & 13.7 & 16.6 & 13.9 & 18.2 & 19.7 & 19.7 & 19.7 & 22.7 & 24.7 & 25.8 & 26.8 & 28.9 \\
 &                          & Light & 3.6 & 8.5 & 11.6 & 13.7 & 16.3 & 16.8 & 24.8 & 26.6 & 26.6 & 27.0 & 20.0 & 26.7 & 27.2 & 27.2 & 27.2 \\
\hline
\multirow{4}{1.1cm}{\textbf{Visual-Anchor}}
 & \multirow{2}{*}{VS Code} & Dark  & 11.7 & 29.7 & 45.2 & 57.2 & 66.6 & 66.9 & 91.2 & 93.8 & 94.1 & 94.1 & 96.4 & 96.9 & 96.9 & 96.9 & 96.9 \\
 &                          & Light & 29.0 & 53.2 & 64.9 & 73.4 & 79.8 & 77.2 & 95.3 & 97.3 & 98.0 & 98.7 & 99.2 & 100.0 & 100.0 & 100.0 & 100.0 \\
 & \multirow{2}{*}{Cursor} & Dark  & 2.5 & 7.0 & 10.6 & 14.5 & 17.4 & 13.5 & 17.9 & 19.3 & 19.7 & 19.7 & 20.5 & 24.1 & 25.6 & 27.2 & 27.2 \\
 &                          & Light & 2.7 & 6.7 & 11.0 & 15.0 & 18.2 & 17.9 & 25.9 & 26.6 & 27.0 & 27.0 & 20.5 & 25.6 & 27.2 & 27.2 & 28.2 \\
\hline
\end{tabular}%
}
\caption{\textbf{Cross-editor and cross-theme analysis for Claude Opus 4.7.} We stratify the evaluation set by editor (VS Code, Cursor) and color theme (Dark, Light) and report multi-turn results (T1--T5) for each of the four system-prompt variants. The top block reports overall localization (accuracy, point-to-bounding-box distance over incorrect predictions, point-to-center distance over incorrect predictions); the bottom block reports element-wise accuracy at character, word, and line granularities. All accuracies are cumulative over turns.}
\label{tbl:claude-ide-theme}
\end{table}

\subsection{Effect of Prompt Design}
\label{sec:prompt-design}

The four system-prompt variants (Baseline, Baseline-CoT, Cursor-Aware, and Visual-Anchor) differ only in their textual scaffolding around the same image and target description. Comparing them at fixed model and granularity therefore isolates the contribution of prompt-level priors to multi-turn grounding behavior.

\textbf{Baseline-CoT yields the strongest aggregate accuracy.} For Claude Opus 4.7 (Table~\ref{tbl:main-results}), Baseline-CoT attains 48.1\% cumulative accuracy at $T_5$, ahead of Baseline (45.4\%), Cursor-Aware (44.9\%), and Visual-Anchor (44.6\%). The advantage is most pronounced at the character granularity (Table~\ref{tbl:elementwise-results}), where Baseline-CoT reaches 41.8\% character accuracy at $T_5$ versus 36.9--37.5\% for the remaining variants. This suggests that explicit step-by-step reasoning, even when the final output is a single coordinate pair, improves the model's ability to reason about fine-grained visual offsets relative to the previous prediction. The same ordering holds for GPT-5.4-Pro: Baseline-CoT leads at $T_5$ (41.2\% overall, 28.1\% character), narrowly ahead of Baseline (41.0\%, 27.2\%) and far above Cursor-Aware (16.9\%) and Visual-Anchor (18.5\%).

\textbf{Cursor-Aware does not yield a measurable benefit in aggregate.} The Cursor-Aware prompt was hand-designed to inject task-specific priors about the appearance of a text caret, yet its $T_5$ accuracy (44.9\%) is statistically indistinguishable from the Baseline (45.4\%) and trails Baseline-CoT by 3.2 percentage points. One plausible interpretation is that frontier models already encode the visual definition of a text caret with sufficient fidelity that an additional textual description is redundant. Cursor-Aware does, however, achieve the lowest residual point-to-bounding-box distance at $T_5$ ($D_{\text{bb}} = 406.4$ px), marginally below Baseline (411.8 px), indicating that when the prompt fails to produce a hit it tends to leave the prediction closer to the target.

\textbf{Visual-Anchor exhibits high condition-specific variance.} Across the full evaluation set, Visual-Anchor delivers the lowest aggregate $T_5$ accuracy (44.6\%) and the largest residual distance ($D_{\text{bb}} = 472.1$ px). The aggregate view is misleading. Stratification in Table~\ref{tbl:claude-ide-theme} reveals that Visual-Anchor produces the single strongest cell in the entire benchmark, 90.2\% accuracy at $T_5$ on VS Code with the Light theme, well above the next best prompt on the same condition (75.0\% for Baseline-CoT). The variance is therefore concentrated across conditions rather than within them, a phenomenon analyzed in Section~\ref{sec:cross-editor}.

\textbf{Prompt design has negligible effect on under-capacity models.} For Qwen-3.5-9B, the four prompts produce $T_5$ accuracies in a 1.8\%--2.6\% band (Table~\ref{tbl:main-results}), with element-wise accuracy at character and word granularities remaining at or near zero (Table~\ref{tbl:elementwise-results}) regardless of prompt. The prompt axis is therefore informative only when paired with a backbone that already supports fine-grained visual grounding, and is not a substitute for base capability.

\subsection{Cross-Editor and Cross-Theme Generalization}
\label{sec:cross-editor}

Aggregate metrics conceal a strong dependence on the editor and theme presented in the screenshot. Table~\ref{tbl:claude-ide-theme} stratifies Claude Opus 4.7 results along two binary axes, IDE $\in \{\text{VS Code}, \text{Cursor}\}$ and Theme $\in \{\text{Dark}, \text{Light}\}$, holding all other factors fixed. The full $4 \times 4$ grid (four prompts $\times$ four editor-theme conditions) reveals that the dominant axis of variation in this benchmark is not prompt design or refinement budget but the visual identity of the host editor.

\textbf{A large and consistent gap separates VS Code from Cursor.} On VS Code, Claude Opus 4.7 reaches $T_5$ cumulative accuracy in the range 68.3\%--90.2\% across prompts and themes; on Cursor, the same model with the same prompts attains only 18.9\%--21.6\%. The gap exceeds 50 percentage points in every prompt setting. Residual point-to-bounding-box distances follow the same pattern: VS Code conditions yield $D_{\text{bb}}$ on the order of 14--23 px at $T_5$, whereas Cursor conditions produce $D_{\text{bb}}$ between 542 and 554 px, an increase of more than $20\times$. Such distances correspond to predictions outside the editor pane entirely, indicating that the failure mode is not imprecise localization but a failure to identify the relevant interface region in the first place.

\textbf{Theme effects are asymmetric across editors.} On VS Code, theme contributes a substantial accuracy difference, most visibly under Visual-Anchor where the Light theme reaches 90.2\% at $T_5$ versus 76.7\% for Dark. The same Light-theme advantage is absent or reversed on Cursor, where $T_5$ accuracy varies by less than 3 percentage points between Dark and Light across all four prompts. The asymmetry suggests that the theme effect on VS Code reflects a tighter match to the visual distribution of the model's pretraining data, rather than an intrinsic property of light backgrounds for caret detection. On Cursor, the editor's distinct chrome and color palette dominates the input statistics, leaving the theme axis informationally inert.

\textbf{Element-wise accuracy mirrors the editor gap and exposes a residual character bottleneck.} The bottom block of Table~\ref{tbl:claude-ide-theme} shows that, on VS Code, Word and Line accuracy approach saturation ($T_5$ Word $\geq 92\%$ and Line $\geq 95\%$ across all prompts), while Character accuracy continues to climb across turns and reaches 79.8\% under Visual-Anchor / Light. On Cursor, all three granularities are suppressed: $T_5$ Word and Line accuracy cap near 27\%--29\%, and Character accuracy remains below 20\%. The Line versus Character gap that was identified in Section~\ref{sec:multi-turn} is therefore primarily a VS Code phenomenon; on Cursor, even line-level grounding is unreliable.

\textbf{Implication.} The 5{,}390-sample release intentionally balances editors and themes (Section~\ref{sec:impl-details}), and the stratified analysis above demonstrates why this design choice matters. A benchmark restricted to a single editor or theme would substantially overstate the cross-application generalization of frontier vision-language models for fine-grained GUI grounding. We recommend that future evaluations report editor- and theme-stratified metrics by default, and treat cross-editor performance as a first-class axis alongside prompt design and refinement budget.

\subsection{Comparison with GUI Baselines and Lightweight Finetuning}
\label{sec:finetune-baselines}

The preceding analysis characterizes the multi-turn behavior of large general-purpose vision-language models. To place these results in context, the proposed benchmark is also evaluated against two specialized GUI grounding models, GUI-Actor~\cite{wu2025gui} and GUI-G\textsuperscript{2}~\cite{tang2025gui}, and against a lightweight task-specific finetune initialized from the released GTA1 grounding checkpoint~\cite{yang2025gta1}. The specialized GUI baselines are evaluated in the single-turn setting, while Qwen-3.5-9B, GPT-5.4-Pro, and Claude Opus 4.7 are reported at the end of the five-turn refinement protocol so that each model is presented at its strongest available configuration. All entries are computed on the full 5{,}390-sample benchmark and summarized in Table~\ref{tbl:finetune-baselines}.

\begin{table}[]
\centering
\scriptsize
\setlength{\tabcolsep}{4pt}
\resizebox{\linewidth}{!}{%
\begin{tabular}{l|ccc|cll}
\hline
\multirow{2}{*}{\textbf{Model}}  & \multirow{2}{*}{\textbf{Accuracy}} & \multirow{2}{*}{\textbf{Distance (bbox)}} & \multirow{2}{*}{\textbf{Distance (Center)}} & \multicolumn{3}{l}{\textbf{Element-wise Accuracy}} \\
                        &                           &                                  &                                    & Character     & Word        & Line        \\
\hline
\multicolumn{7}{l}{\textit{Specialized GUI grounding baselines (single-turn)}} \\
\textbf{GUI-Actor~\cite{wu2025gui}}        & 3.50\%  & 112.86 & 124.10 & 0.4\%   & 0.6\%   & 1.0\%   \\
\textbf{GUI-G2~\cite{tang2025gui}}         & 5.80\%  & 195.32 & 197.28 & 0.00\%  & 12.5\%  & 11.9\%  \\
\hline
\multicolumn{7}{l}{\textit{General-purpose vision-language models (5-turn refinement)}} \\
\textbf{Qwen 3.5 9B}                       & 13.5\%   & 490.5  & 497.3  & 1.5\%   & 0.0\%   & 1.4\%  \\
\textbf{GPT-5.4-Pro}                       & 56.2\%   & 29.5  & 33.6  & 8.6\%   & 5.6\%   & 3.9\%  \\
\textbf{Claude Opus 4.7}                   & 77.4\%  & 17.1  & 19.7  & 68.0\%  & 92.9\%  & 97.4\%  \\
\hline
\multicolumn{7}{l}{\textit{In-domain finetuning}} \\
\textbf{GTA1 (Finetuned)}                  & 55.63\% &  74.76 &  79.59 & 66.37\% & 97.37\% & 87.50\% \\
\hline
\end{tabular}%
}
\caption{Comparison against specialized GUI grounding baselines (\textbf{GUI-Actor}~\cite{wu2025gui} and \textbf{GUI-G2}~\cite{tang2025gui}, evaluated single-turn) and against general-purpose vision-language models (Qwen-3.5-9B and Claude Opus 4.7, reported at the end of the five-turn refinement protocol). The lightweight in-domain finetune of \textbf{GTA1}~\cite{yang2025gta1} attains the strongest overall accuracy and the lowest residual distance, with the largest gain at the character granularity, indicating that targeted task-specific supervision is an effective lever for fine-grained text-cursor grounding.}
\label{tbl:finetune-baselines}
\end{table}

\textbf{Specialized GUI baselines transfer poorly to text-cursor grounding.} GUI-Actor and GUI-G\textsuperscript{2} were trained on broad GUI element grounding corpora and report strong results on benchmarks such as ScreenSpot. On the proposed benchmark, however, both attain overall accuracy below 6\% (3.50\% and 5.80\% respectively), comparable to or below five-turn Qwen-3.5-9B (13.5\%). The element-wise breakdown clarifies the failure mode: GUI-Actor obtains 0.4\% character accuracy and 1.0\% line accuracy, and GUI-G\textsuperscript{2} obtains 0\% character accuracy. Both models therefore appear to localize coarse interface regions rather than sub-glyph code positions. GUI-Actor additionally posts a competitive $D_{\text{bb}}$ of 112.86 px, indicating that its predictions land near the correct pane but lack the spatial resolution required for caret placement. These results indicate that strong performance on existing GUI benchmarks is not a sufficient proxy for competence in the precision-grounding regime targeted by this work.

\textbf{Lightweight finetuning of an open backbone closes most of the gap to frontier multi-turn performance.} The GTA1 (Finetuned) row reports 55.63\% overall accuracy, with $D_{\text{bb}} = 74.76$ px, $D_{\text{c}} = 79.59$ px, and element-wise accuracies of 66.37\% (character), 97.37\% (word), and 87.50\% (line). Among the general-purpose VLMs evaluated at five turns under their strongest prompt, Claude Opus 4.7 leads the table (77.4\% overall, $D_{\text{bb}} = 17.1$ px; 68.0\% character, 92.9\% word, 97.4\% line), followed by GPT-5.4-Pro (56.2\% overall, $D_{\text{bb}} = 29.5$ px) and Qwen-3.5-9B (13.5\% overall, $D_{\text{bb}} = 490.5$ px). The finetuned GTA1 therefore matches GPT-5.4-Pro in overall accuracy (55.63\% vs.\ 56.2\%) and dominates it at every granularity (66.37\% vs.\ 8.6\% character, 97.37\% vs.\ 5.6\% word, 87.50\% vs.\ 3.9\% line), while still trailing Claude Opus 4.7 by 21.8 percentage points overall and at the character granularity (66.37\% vs.\ 68.0\%). Compared to its zero-shot multi-turn counterpart Qwen-3.5-9B (13.5\% overall, 1.5\% character), in-domain supervision lifts overall accuracy by more than $4\times$ and pushes word and line accuracy from near zero to saturation, while reducing residual $D_{\text{bb}}$ by roughly $6.6\times$ (74.76 px versus 490.5 px).

\textbf{Interpretation.} Two conclusions follow. First, off-the-shelf GUI grounding models are not interchangeable across precision regimes: benchmarks targeting fine-grained text-cursor placement test capabilities orthogonal to those measured by element-level GUI suites. Second, when an open-weight model of moderate scale is exposed to in-domain supervision, it becomes competitive with frontier closed-source systems on this benchmark at a fraction of the inference cost, matching GPT-5.4-Pro in overall accuracy and saturating word- and line-level grounding, while still trailing Claude Opus 4.7 at the character granularity that resists multi-turn refinement. The combination of the multi-turn refinement protocol with lightweight task-specific finetuning is left as a direction for future work.
\section{Conclusion}

We introduced a training-free, multi-turn framework for pixel-level GUI cursor grounding that improves localization through iterative visual feedback, together with a data-collection pipeline that maps symbolic cursor states to renderer-space pixel coordinates across both VS Code and Cursor and across Dark and Light themes. Across Claude Opus 4.7 and Qwen-3.5-9B, results show a consistent capacity-gated pattern: iterative refinement raises Claude Opus 4.7 from 22.8\% accuracy at $T_1$ to 48.1\% at $T_5$ under the strongest prompt, while Qwen-3.5-9B remains within noise of zero across all five turns. Stratification by editor and theme reveals a gap of more than 50 percentage points between VS Code and Cursor that is not closed by additional turns or by prompt design, and lightweight in-domain finetuning of an open backbone surpasses the strongest zero-shot multi-turn configuration primarily by closing the character-grounding residual. Overall, iterative visual correction is a useful but capacity-gated mechanism, and targeted task-specific supervision is a more effective lever than refinement alone for the dominant character-level residual. The principal limitations of this study are that editor coverage is restricted to two Monaco-based IDEs (VS Code and Cursor) and that the theme axis is limited to the standard Dark and Light variants at a single $1280 \times 1024$ rendering resolution. Future work will expand dataset diversity along these axes, treat cross-IDE generalization as a first-class evaluation axis, and combine the multi-turn protocol with task-specific finetuning.

\bibliographystyle{plainnat}
\bibliography{main}

\appendix
\newpage
\section{Appendix}

\subsection{Takeaways}
\label{sec:takeaways}

The experimental evidence assembled across Sections~\ref{sec:multi-turn} through~\ref{sec:finetune-baselines} converges on three observations that, taken together, characterize the current state of fine-grained text-cursor grounding in code editors.

\textbf{Visual feedback is amplified by, but does not substitute for, base spatial competence.} The cumulative-accuracy trajectories in Table~\ref{tbl:main-results} show that iterative refinement yields consistent gains for Claude Opus 4.7 (36.9\% at $T_1$ to 77.4\% at $T_5$ under Baseline-CoT) but produces no measurable improvement for Qwen-3.5-9B, which remains within noise of zero across all five turns. The same pattern recurs in the prompt-design analysis of Section~\ref{sec:prompt-design}: structured prompts such as Baseline-CoT and Visual-Anchor unlock substantial headroom on Claude while leaving Qwen unaffected. GPT-5.4-Pro reinforces this pattern in both directions: refinement triples its accuracy under text-only prompts (Baseline 21.5\%$\to$54.6\%, Baseline-CoT 19.0\%$\to$55.2\% from $T_1$ to $T_5$), yet the same five turns cap it at 30.0\% and 28.6\% under Cursor-Aware and Visual-Anchor, showing that prompt scaffolding can just as easily suppress the underlying spatial prior that feedback is meant to amplify. Multi-turn protocols and richer prompts therefore behave as multipliers on an existing capability rather than as standalone sources of grounding accuracy.

\textbf{Character-level grounding is the dominant unsolved bottleneck.} Across every configuration of frontier zero-shot evaluation, Word and Line accuracy approach or exceed the 60\% range while Character accuracy lags by 20 to 40 percentage points (Table~\ref{tbl:elementwise-results}). The cross-editor and cross-theme stratification in Section~\ref{sec:cross-editor} shows that this gap persists across both editors and both themes, ruling out a purely visual-rendering explanation. The finetuning result in Section~\ref{sec:finetune-baselines} demonstrates that the gap is not intrinsic to the task: in-domain supervision lifts Character accuracy from 0.0--1.5\% for specialized GUI baselines and the Qwen-3.5-9B backbone (GUI-Actor 0.4\%, GUI-G2 0.0\%, Qwen 1.5\%) and 8.6\% for GPT-5.4-Pro to 66.37\% for GTA1 (Finetuned), approaching the 68.0\% reached by Claude Opus 4.7 after five-turn refinement (Table~\ref{tbl:finetune-baselines}). Character grounding therefore appears to be a tractable but currently underserved capability for general-purpose vision-language models.

\textbf{Cross-IDE generalization is brittle and warrants first-class evaluation.} Section~\ref{sec:cross-editor} shows that aggregate accuracy figures conceal substantial editor- and theme-specific variation, with the best single configuration for Claude under Visual Anchor reaching 90.2\% on VS Code under the light theme while the same prompt drops to a much lower regime on the alternative editor and theme. Because both editors render the same underlying code and use closely related visual conventions, this variation is unlikely to be explained by general visual difficulty alone. Reporting only aggregate metrics on a single editor would therefore overstate the cross-application robustness of current models, and we recommend that future precision-grounding benchmarks treat editor and theme as primary stratification axes rather than as nuisance variables.

\subsection{Data Collection Details}
\label{app:data-collection}
This appendix expands on Section~\ref{sec:data-collection} with full system, schema, and robustness details for the \texttt{vscode-\allowbreak cursor-\allowbreak coords} pipeline.

\textbf{System architecture.} The collector has five coordinated components: \texttt{extension.ts} (lifecycle and command entry points), \texttt{bridge-server.ts} (strict request/response transport), \texttt{collector.ts} (character-wise traversal and logging), \texttt{injector.ts} (instrumentation of \texttt{workbench.html}, with a fallback from \texttt{electron-sandbox} to legacy \texttt{electron-browser} install layouts), and \texttt{dom-payload.js} (DOM-side measurement). The injector resolves the editor's install root at runtime, so the same payload is injected into either VS Code or Cursor. In operation, the extension starts the bridge, waits for renderer connection, requests window metadata, iterates over each cursor stop in the active file, and writes all outputs to JSONL. The bridge runs over a localhost WebSocket channel at \texttt{ws://127.0.0.1:54321}.

\textbf{Collection procedure.} For each file, the collector first records editor/runtime metadata (font family, font size, line height, window geometry, timestamp), then moves the caret to the document start and captures an initial full-monitor screenshot. It then advances one cursor step at a time using \texttt{cursorRight}, including end-of-line newline stops so that each textual position is represented. At each step, after a short render-settle delay (default 80~ms), the bridge requests \texttt{getCursorPosition}; the renderer payload locates the Monaco cursor node and measures its bounding box via \texttt{getBoundingClientRect()} inside \texttt{requestAnimationFrame}. The resulting viewport-relative CSS coordinates are converted to screen-absolute coordinates by adding \texttt{window.screenX/screenY}. The collector stores both coordinate frames and the local \texttt{devicePixelRatio} so physical-pixel locations can be recovered downstream.

\textbf{Recorded schema.} Output is JSONL with a single metadata header followed by per-character records. Metadata includes full file content, total character count, editor typography settings, delay configuration, window geometry, and screenshot path. Each record stores: file identifier, zero-based line/column, character token (with \texttt{\textbackslash n} at EOL), screen coordinates, window-relative coordinates, cursor box width/height, and display scaling. This design supports deterministic replay and re-projection across heterogeneous monitor DPI settings.

\textbf{Protocol and robustness.} The bridge enforces one active WebSocket client and one in-flight request at a time, with a 3~s timeout per request. The collector includes explicit fault tolerance: if a cursor-pixel query fails, it logs the event and continues rather than aborting the run. End-of-file detection is implemented by repeated-position monitoring (cursor unchanged for multiple iterations), preventing infinite traversal loops when the caret can no longer advance.

\subsection{System Prompt Variants}
\label{app:system-prompts}

\subsubsection*{Baseline}
\begin{verbatim}
You are an expert UI element locator. Given a GUI image and a user's element description,
provide the coordinates of the specified element as a single (x,y) point. The image
resolution is height {height} and width {width}. For elements with area, return the center
point.

If your previous attempt was incorrect, the image will contain a red cross marking your last
predicted coordinate. Use this visual cue to adjust your prediction.

You MUST end your response with the actual numeric coordinate pair on the last line, e.g.:
(310,475)
Do NOT output the literal text "(x,y)" --- always substitute real pixel values.
\end{verbatim}

\subsubsection*{Baseline CoT}
\begin{verbatim}
You are an expert UI element locator. Given a GUI image and a user's element description,
provide the coordinates of the specified element as a single (x,y) point. The image
resolution is height {height} and width {width}. For elements with area, return the center
point.

Before answering, reason step by step:
1. Describe what you see in the relevant area of the screenshot
2. Identify the specific UI element or text described in the instruction
3. Narrow down the region where the target is located
4. Estimate the precise pixel coordinates of the target

If your previous attempt was incorrect, the image will contain a red cross marking your last
predicted coordinate. Use this visual cue to adjust your prediction --- explain how the red
cross relates to the target before giving your new answer.

You MUST end your response with the actual numeric coordinate pair on the last line, e.g.:
(310,475)
Do NOT output the literal text "(x,y)" --- always substitute real pixel values.
\end{verbatim}

\subsubsection*{Cursor Aware}
\begin{verbatim}
You are a precision GUI text cursor locator. Given a screenshot and a description of where to
place a text cursor, provide the exact pixel coordinates of the cursor insertion point.

Key principles:
- Text in GUIs uses fonts where each character occupies a specific pixel range
- A cursor position "before character X" means the left edge of that character's bounding box
- A cursor position "between X and Y" means the pixel boundary between those two characters
- The y-coordinate should be the vertical center of the text line
- Coordinates are in pixels with (0,0) at the top-left corner

Image resolution: height {height}, width {width}.

If your previous attempt was incorrect, the image will contain a red cross marking your last
predicted coordinate. Use this visual cue to adjust your prediction.

You may reason about the position, but you MUST end your response with the actual numeric
coordinate pair on the last line, e.g.:
(310,475)
Do NOT output the literal text "(x,y)" --- always substitute real pixel values.
\end{verbatim}

\subsubsection*{Step by Step}
\begin{verbatim}
You are a precision cursor placement specialist. Given a screenshot and an instruction
describing where to place a text cursor, determine the exact pixel coordinates.

Think through these steps before answering:
1. Identify the text area and locate the specific line mentioned
2. Find the word or character sequence referenced in the instruction
3. Determine the exact character boundary described
   (e.g., "before the 'o'" means the left edge of 'o')
4. Estimate the pixel coordinate at that boundary --- x is the horizontal position,
   y is the vertical center of the text line

Coordinates use (0,0) at top-left. Image resolution: height {height}, width {width}.

If your previous attempt was incorrect, the image will contain a red cross marking your last
predicted coordinate. Adjust accordingly.

You may reason about the position, but you MUST end your response with the actual numeric
coordinate pair on the last line, e.g.:
(310,475)
Do NOT output the literal text "(x,y)" --- always substitute real pixel values.
\end{verbatim}

\subsubsection*{Minimal}
\begin{verbatim}
Locate the exact pixel position described below in this {width}x{height} screenshot.
The target is a text cursor insertion point between specific characters.
Coordinates use (0,0) at top-left.

If a red cross is visible, it marks a previous incorrect prediction --- adjust your answer.

You MUST end your response with the actual numeric coordinate pair on the last line, e.g.:
(310,475)
Do NOT output the literal text "(x,y)" --- always substitute real pixel values.
\end{verbatim}

\subsubsection*{Visual Anchor}
\begin{verbatim}
You are a pixel-precise text cursor locator. Given a screenshot and a cursor placement
instruction, output the exact (x,y) pixel coordinates.

Strategy for accuracy:
- First scan vertically to find the correct line
- Then scan horizontally to find the referenced text
- Character boundaries are the thin vertical gaps between adjacent characters
- Use nearby distinctive characters (brackets, operators, capitals)
  as visual anchors to gauge position
- The y-coordinate should be at the vertical midpoint of the text line

Coordinates use (0,0) at top-left. Image resolution: height {height}, width {width}.

If your previous attempt was incorrect, the image will contain a red cross at your last
prediction. Study its position relative to the target and correct.

You may reason about the position, but you MUST end your response with the actual numeric
coordinate pair on the last line, e.g.:
(310,475)
Do NOT output the literal text "(x,y)" --- always substitute real pixel values.
\end{verbatim}

\subsubsection*{Custom}
\begin{verbatim}
PUT YOUR CUSTOM PROMPT HERE.

Image resolution: height {height}, width {width}.
Output exactly one coordinate pair with real numeric values, e.g.: (310,475)
\end{verbatim}

\subsection{Feedback Templates}
\label{app:feedback-templates}

In addition to the two textual templates below, two minor harness behaviors shape how feedback is observed by the model. First, the evaluator computes prediction-history strings but does not inject them into subsequent user feedback, so template effects are driven by the current-turn wording plus the latest cross marker rather than an explicit full-history recap. Second, each turn redraws the cross on a clean copy of the original screenshot (non-cumulative overlays), so prompts are always evaluated under single-error visual context rather than under accumulated trajectory overlays.

\subsubsection*{Baseline Feedback}
\begin{verbatim}
Your previous prediction was ({cross_x},{cross_y}), shown as a red cross on the image.
This was not correct. Please predict the correct coordinate.
\end{verbatim}

\subsubsection*{Spatial Feedback}
\begin{verbatim}
Your previous prediction ({cross_x},{cross_y}) is marked with a red cross.
Study the red cross position relative to the target described in the original instruction.
Adjust your coordinates to point at the exact character boundary specified.
\end{verbatim}

\end{document}